\documentclass[a4paper,twocolumn,11pt]{article}

\usepackage[english]{babel}
\usepackage[a4paper,left=1cm,right=1cm,top=2.5cm,bottom=2.5cm]{geometry}
\usepackage{amsmath}
\usepackage{amsfonts}
\usepackage{amssymb}
\usepackage{mathtools}
\usepackage{hyperref}
\usepackage{csquotes}
\usepackage[dvipsnames]{color}
\usepackage{subcaption}
\usepackage{tikz-cd}
\usepackage{enumitem}
\usepackage{float}
\usepackage{stfloats}

\usepackage{hyperref}
\hypersetup{colorlinks=true,linkcolor=blue,urlcolor=blue}

\usepackage{graphicx}
\graphicspath{{./images/}}

\usepackage{titlesec}
\titlelabel{\thetitle.\ }
\titleformat*{\section}{\Large\bfseries\sffamily}
\titleformat*{\subsection}{\large\bfseries\sffamily}

\usepackage{abstract}

\usepackage{amsthm}
\newtheoremstyle{mytheoremstyle}{7pt}{7pt}{\normalfont}{}{\normalfont\bfseries}{:}{.5em}{}
\theoremstyle{mytheoremstyle}
\newtheorem{definition}{Definition}
\newtheorem{proposition}{Proposition}
\newtheorem{remark}{Remark}
\newtheorem*{myproof}{Proof}

\newcommand{\keyword}[1]{\textsl{#1}}

\newcommand{\setR}{\mathbb{R}}
\DeclareMathOperator{\Tr}{Tr}

\DeclareMathOperator{\Exp}{\mathbb{E}}

\DeclareMathOperator{\Enc}{\mathsf{E}}
\DeclareMathOperator{\Dec}{\mathsf{D}}

\begin{document}

\title{\bfseries\sffamily Assessing local deformation and computing scalar curvature with nonlinear conformal regularization of decoders}

\author{
    Benjamin \textsc{Couéraud}\footnote{Corresponding author, please address all communications to \texttt{bcoueraud87@gmail.com}.}\\
    Zuse Institute Berlin
    \and
    Vikram \textsc{Sunkara}\\
    Zuse Institute Berlin
    \and
    Christof \textsc{Sch\"utte}\\
    Zuse Institute Berlin
}

\maketitle

\begin{abstract}
    One aim of dimensionality reduction is to discover the main factors that explain the data, and as such is paramount to many applications. When working with high dimensional data, autoencoders offer a simple yet effective approach to learn low-dimensional representations. The two components of a general autoencoder consist first of an encoder that maps the observed data onto a latent space; and second a decoder that maps the latent space back to the original observation space, which allows to learn a low-dimensional manifold representation of the original data.   In this article, we introduce a new type of geometric regularization  for decoding maps approximated by deep neural networks, namely \keyword{nonlinear conformal regularization}. This regularization procedure permits local variations of the decoder map and comes with a new scalar field called \keyword{conformal factor} which acts as a quantitative indicator of the amount of local deformation sustained by the latent space when mapped into the original data space. We also show that this regularization technique allows the computation of the scalar curvature of the learned manifold. Implementation and experiments on the Swiss roll and CelebA datasets are performed to illustrate how to obtain these quantities from the architecture.
\end{abstract}



\section{Introduction}

Dimensionality reduction, also known as manifold learning in certain contexts, aims at representing high-dimensional datasets by a smaller number of latent factors. Principal Component Analysis (PCA), is a well-known method that achieves that aim, but is not satisfactory for highly nonlinear datasets. Starting approximately two decades ago, numerous methods have been developed to address nonlinearity in dimension reduction. Popular among these methods are Locally Linear Embedding (LLE) \cite{RS00}, Multidimensional Distance Scaling (MDS) \cite{Tor52}, Laplacian eigenmaps \cite{BN01}, just to name a few. These methods rely on performing a Singular Value Decomposition (SVD) at some stage and as such are not directly translatable to large and high-dimensional datasets. For an in-depth overview of nonlinear dimensionality reduction, see~\cite{review}. In the last decade, deep neural networks (DNNs) have been used in the analysis of such highly nonlinear datasets with much success. DNNs help in circumventing the need for SVD by optimizing with respect to a reconstruction loss (possibly amended with regularizers), on randomly selected points, to learn both local and global properties of the data. 

We begin by establishing the mathematical notations used in this article. Let us denote the \keyword{observation (data) space} as $X\subset\setR^n$, with $n$ usually very large, representing the number of features of the dataset. After performing dimension reduction via an \keyword{encoder} map $\Enc$, we end up in the \keyword{latent space} $Z\subset\setR^m$, with $m\ll n$. To reconstruct this reduced data, we map it back with a \keyword{decoder} map $\Dec$ to the original space $X$, which is usually denoted by $\widehat{X}$ to emphasize the reconstruction process. Therefore, dimensionality reduction is represented by the maps:
\begin{equation}
    x\in X\subset\setR^n\xmapsto{\quad\Enc\quad}z\in Z\subset\setR^m\xmapsto{\quad\Dec\quad}\hat{x}\in\widehat{X}\subset\setR^n.\label{eq:autoencoder}
\end{equation}
The manifold that is learned is $\mathcal{D}=\Dec(Z)\subset\setR^n$; it can be seen as embedded in $\setR^n$ although the existence of an embedding (in the mathematical sense) is not guaranteed. In practice, the latent space $Z$ and the decoder $\Dec$ constitute a \keyword{parametrization} of the learned manifold $\mathcal{D}$. When the encoder $\Enc$ and decoder $\Dec$ are neural networks, the structure above is knows as an \keyword{autoencoder}. In this context the encoder is parameterized by weights $\theta_\mathsf{enc}$ and the decoder is parameterized by weights $\theta_\mathsf{dec}$, and the optimal weights are found by minimizing the \keyword{reconstruction loss}
\begin{equation}
    \label{eq:reconstruction-loss}
    \mathcal{L}_{\mathsf{recon}}(\Enc,\Dec)=\frac{1}{N}\sum_{i=1}^N\Big\|x_i-\Dec\big(\Enc(x_i;\theta_\mathsf{enc});\theta_\mathsf{dec}\big)\Big\|^2_2,
\end{equation}
with respect to the weights $\theta_\mathsf{enc}$ and $\theta_\mathsf{dec}$, and where $x_i$ denotes one of the $N$ data samples ($\|\cdot\|_2$ denotes the Euclidean norm). In what follows we will omit the reference to these weights for simplicity.

While using the reconstruction loss guarantees that the data manifold $\mathcal{D}$ is close to the original data found in $X$, neighboring data points in $X$ can me mapped to points in $Z$ (called \keyword{codes}) which are far away from each other in the latent space $Z$, and similarly nearby codes in $Z$ can be reconstructed without preserving the distance between them, resulting in encoders and decoders that exhibit high variations. In order to obtain better reduced representations of the data in the latent space, various \keyword{geometric} regularization terms have been proposed for either the encoder or the decoder, in order to not tear apart the latent representation and preserve neighborhoods (topology), or the distances in a certain sense (geometry -- which practitioners may want to use to make statements about their datasets)\footnote{Regularizers can be seen as \keyword{constraints} in a given optimization problem and serve many purposes, such as imposing sparsity or smoothness. In this article regularization is used to enforce geometric properties.}. For instance, to avoid high variations of the encoder, the norm of the Jacobian of the encoder is used as a regularizer (see \cite{ContractiveAE}), which encourages the variations to be \keyword{globally} small. In another direction, a \keyword{global isometry} can be encouraged for the decoder by using a regularization term of the form:
\begin{align}
    \begin{split}
        \mathcal{L}_{\mathsf{globiso}}&(\Dec)\\
        &=\sum_{z_1,z_2\in Z}\Big|d_Z(z_1,z_2)-d_X(\Dec(z_1),\Dec(z_2))\Big|,
    \label{eq:global-isometric-regularizer}
    \end{split}
\end{align}
where $d_Z$ denotes a distance on the latent space $Z$ and $d_X$ denotes a distance on the data space $X$ (or similarly for the encoder; see \cite{DIMAL} for an application to encoders where $d_X$ is the (approximated) \keyword{geodesic distance}). However, this regularization term is strong, as distances are encouraged to be preserved for \keyword{any} pair of codes, possibly not near each other. A weaker possibility is to encourage the decoder to be a \keyword{local isometry}, meaning that distances are preserved only between mapped neighborhoods. If both the data space and latent space are equipped with Euclidean metrics, the regularizer for the decoder is then 
\begin{equation}
    \label{eq:local-isometric-regularizer-ninv}
    \mathcal{L}_{\mathsf{lociso}}(\Dec)=\sum_{z\in Z}\Big\|J_{\Dec}(z)^\mathsf{T}J_{\Dec}(z)-I_m\Big\|_F,
\end{equation}
where $J_{\Dec}(z)$ denotes the Jacobian of the decoder at a point $z\in Z$ (see \cite{IsometricAE} for such an approach), $I_m$ the identity matrix of size $m\times m$, and $\|\cdot\|_F$ the Frobenius norm. Local distance preservation has also been explored in \cite{DMT}, without resorting to Jacobians but by using an auxiliary neighborhood graph and enforcing a local version of \eqref{eq:global-isometric-regularizer} between consecutive layers of the encoder. Another direction for geometric regularization makes use of curvature minimization, see for instance \cite{RegCurvature}.

\section{Nonlinear conformal decoding}
\label{sec:conformal-decoding}

In \cite{lee2023geometricperspectiveautoencoders}, it was noted that \eqref{eq:local-isometric-regularizer-ninv} is not a coordinate-invariant expression. The authors then proposed a new coordinate-invariant regularizer that ensures that the learned decoder is as closed as possible to a (local) isometry, and extended this approach (in a slightly different way) to conformal mappings possessing a \keyword{constant} conformal factor (see also \cite{FMVAE} and \cite{LOCA} for former approaches). Conformal maps locally preserve distances up to a (possibly nonlinear) factor, and therefore preserve angles. Such maps are well-known in cartography (for cartographic projections, akin to what an encoder does) as well as physics and engineering (for reformulating problems into equivalent yet simpler ones). In generative artificial intelligence, conformal maps have also been used as normalizing flows (see \cite{ConformalGenAI}). We extend this approach to the case a \keyword{nonlinear conformal factor} and show its benefits.

\begin{definition}
    Let $(M,g)$ and $(N,h)$ be two Riemannian manifolds and $f:M\to N$ be a smooth map. The map $f$ is called \keyword{conformal} if there exists a smooth map $c:M\to\mathbb{R}$ everywhere strictly positive such that $f^\star h=cg$, which means that for any point $x\in M$, $u$, $v\in T_xM$, we have: 
    $$
        h_{f(x)}\big(\mathrm{d}f_x(u),\mathrm{d}f_x(v)\big)=c(x)g_x(u,v),
    $$
    or alternatively in terms of the corresponding matrices:
    \begin{equation}
        J_f(x)^\mathsf{T}H\big(f(x)\big)J_f(x)=c(x)G(x).
        \label{eq:conformal-jacobians}
    \end{equation}
\end{definition}
The function $c$ (or sometimes $\sqrt{c}$) is known as the \keyword{conformal factor}).

We now would like to encourage a decoder $\Dec:Z\to\widehat{X}$ to be a \keyword{nonlinear} conformal map. In many practical cases, the Riemannian metric on the data space $X$ (or $\widehat{X}$) is assumed to be Euclidean, and we will assume so as well: it amounts to say that the matrix $H$ above is everywhere the identity matrix. Then from \eqref{eq:conformal-jacobians}, the condition we are trying to enforce reads $J_{\Dec}(z)^\mathsf{T}J_{\Dec}(z)G(z)^{-1}=c(z)I_m$, where $G(z)$ is the matrix of the Riemannian metric on the latent space $Z$ (which is invertible) and $c(z)$ is the conformal factor for the decoder $\Dec$. Following \cite{lee2023geometricperspectiveautoencoders}, since eigenvalues of a matrix are invariant upon a coordinate change, we need to formulate a regularization term that involves eigenvalues of the ratio matrix $R(z):=J_{\Dec}(z)^\mathsf{T}J_{\Dec}(z)G(z)^{-1}$, which we will denote by $\lambda_i(z)$, $0\leq i\leq m$. A natural expression that doesn't favor any eigenvalue is:
\begin{equation}
    \label{eq:isometric-regularizer}
    \mathcal{L}_{\mathsf{lociso}}(\Dec)=\frac{1}{m}\sum_{i=1}^m\int_Z\Phi(\lambda_i(z))\,\mathrm{d}\nu,
\end{equation}
where $\Phi:\setR\to\setR$ denotes a smooth, positive and convex function that reaches its minimum at $1$, and $\nu$ denotes a probability measure on the latent space $Z$\footnote{The latent space will always be assumed to be compact.}. This expression can be used to encourage the decoder to be a local isometry; we will explain how briefly hereafter. However, for the purpose of encouraging the decoder $\Dec$ to be a \keyword{nonlinear} conformal map, we propose the following regularization term:
\begin{equation}
    \label{eq:nonlinear-conformal-regularizer}
    \widetilde{\mathcal{L}}_{\mathsf{conf}}(\Dec)=\frac{1}{m}\sum_{i=1}^m\int_Z\Phi\left(\frac{\lambda_i(z)}{\sigma(\lambda_1(z),\dots,\lambda_m(z))}\right)\,\mathrm{d}\nu,
\end{equation}
where $\sigma:\setR^m\to\setR$ denotes any symmetric function such that $\sigma(1,\dots,1)=1$ and such that $\sigma$ is homogeneous of degree $1$, meaning $\sigma(\alpha x_1,\cdots,\alpha x_m)=\alpha\sigma(x_1,\dots,x_m)$ for any nonzero real number $\alpha$. Note that such a function $\sigma$ may still favor certain eigenvalues, but as \textit{a priori} there is no reason to do so, we will specify a neutral $\sigma$ later on. This new regularization term is a direct generalization of:
\begin{equation}
    \label{eq:conformal-regularizer}
    \mathcal{L}_{\mathsf{conf}}(\Dec)=\frac{1}{m}\sum_{i=1}^m\int_Z\Phi\left(\frac{1}{I}\lambda_i(z)\right)\,\mathrm{d}\nu,
\end{equation}
with $I=\int_Z\sigma(\lambda_1(z),\dots,\lambda_m(z))\ \mathrm{d}z$ (see \cite[section 3.3]{lee2023geometricperspectiveautoencoders}), with the advantage that we will have access to a \keyword{statistical estimator} of the \keyword{local} deformation factor, see \eqref{eq:nonlinear-conformal-probabilistic-regularizer}. Indeed, the conformal factor at a point $z$ can be seen as the ratio of the area of a ball centered at $z$ to the area of the preimage by the decoder of the same ball, giving a quantitative way to measure the stretching sustained while decoding. 

\begin{proposition}
    Given a decoder $\Dec:Z\to\widehat{X}$, a function $\Phi:\setR\to\setR$ and a function $\sigma:\setR^m\to\setR$ satisfying the properties mentioned above, we have:
    \begin{enumerate}[label*=\protect\fbox{\arabic{enumi}}]
        \item $\widetilde{\mathcal{L}}_{\mathsf{conf}}(\Dec)\geq 0$, 
        \item $\widetilde{\mathcal{L}}_{\mathsf{conf}}(\Dec)=0$ is equivalent to the existence of a smooth map $c:Z\to\setR$ such that $\lambda_i(z)=c(z)$ for all $0\leq i\leq m$.
    \end{enumerate}
    Also, given two decoders $\Dec_1:Z\to\widehat{X}$ and $\Dec_2:Z\to\widehat{X}$, we have:
    \begin{enumerate}[label*=\protect\fbox{\arabic{enumi}}]
        \setcounter{enumi}{2}
        \item If there exists a smooth map $c:Z\to\setR$ such that $R_1(z)=c(z)R_2(z)$, then $\widetilde{\mathcal{L}}_{\mathsf{conf}}(\Dec_1)=\widetilde{\mathcal{L}}_{\mathsf{conf}}(\Dec_2)$.
    \end{enumerate}
\end{proposition}

\begin{myproof}
    The first point is trivial. Concerning the second point, the implication from right to left is clear thanks to the properties of $\sigma$. The converse comes from the fact that $h$ is positive, so $$\Phi\left(\frac{\lambda_i(z)}{\sigma(\lambda_1(z),\dots,\lambda_m(z))}\right)=0$$ for all $0\leq i\leq m$, but then $\Phi$ reaches its unique minimum at $1$, hence setting $c(z):=\sigma(\lambda_1(z),\dots,\lambda_m(z))$ we obtain $\lambda_i(z)=c(z)$ for all $0\leq i\leq m$. The third point is clear using the homogeneity of $\sigma$ after remarking that the eigenvalues of the matrix $R_1$ are the ones of $R_2$ multiplied by the conformal factor $c$, at any point $z$.\hfill $\Box$
\end{myproof}

The first property ensures that the minimum of the regularizer is zero, while the second ensures that this minimum is reached precisely when the decoder is a conformal map, with the conformal factor given by the map $c:Z\to\setR$. The third property simply says that conformal-equivalent decoders yield the same value of the regularization term; such property reduces the search space.

\begin{proposition}
    \label{pro:conformal-probabilistic-regularizer}
    Choosing $\Phi(x)=\frac{1}{2}(x-1)^2$ and $\sigma(x_1,\cdots,x_m):=\frac{1}{m}(x_1+\cdots+x_m)$ so as not to favor any particular eigenvalue, we obtain:
    \begin{gather}
        \label{eq:nonlinear-conformal-probabilistic-regularizer}
        \widetilde{\mathcal{L}}_{\mathsf{conf}}(\Dec)=\frac{m}{2}\Exp_{z\sim\nu}\left[\frac{\Tr R^2(z)}{(\Tr R(z))^2}\right]-\frac{1}{2},\\
        \label{eq:conformal-factor-formula}
        c(z)=\frac{1}{m}\Tr R(z).
    \end{gather}
\end{proposition}

\begin{myproof}
    From the previous proof, we note that the conformal factor is given by $c(z)=\sigma(\lambda_1(z),\dots,\lambda_m(z))=\frac{1}{m}\sum_{i=1}^m\lambda_i(z)$, hence we obtain the second formula. For the first, after rewriting the integral in \eqref{eq:nonlinear-conformal-regularizer} as an expectation with respect to the probability distribution $\nu$ on $Z$, we compute:
    \begin{equation*}
        \begin{split}
            \widetilde{\mathcal{L}}_{\mathrm{conf}}(\Dec)=\frac{1}{2m}\Exp_{z\sim\nu}\Bigg[\sum_{i=1}^m& \frac{m^2\lambda_i(z)^2}{\big(\sum_{j=1}^m\lambda_j(z)\big)^2}\\
            &-\sum_{i=1}^m\frac{2m\lambda_i(z)}{\sum_{j=1}^m\lambda_j(z)}+m\Bigg],
        \end{split}
    \end{equation*}
    which after simplification yields the first formula, since thanks to the invariance by similarity (change of basis) property of the trace we have $\Tr R^2(z)=\sum_{i=1}^m\lambda_i(z)^2$.\hfill $\Box$
\end{myproof}

\begin{remark}
    If we denote by $d_{\mathsf{geo}}^X$ the geodesic distance on the data space $X$ and the geodesic distance on the latent space $Z$ by $d_{\mathsf{geo}}^Z$, we have:
    $$
        d_{\mathsf{geo}}^X\big(\Dec(z_1),\Dec(z_2)\big)\approx c(z)d_{\mathsf{geo}}^Z(z_1,z_2),
    $$
    for any $z_1$, $z_2\in Z$ in a neighborhood of a common $z\in Z$. Indeed, using the Riemannian exponential map $\exp:T_{\Dec(z)}X\to X$ at the point $\Dec(z)$ we have $d_{\mathsf{geo}}^X(\Dec(z_1),\Dec(z_2))=d_{\mathsf{geo}}^X(\mathrm{e}^{\mathrm{d}\Dec_{z_1}(u_1)},\mathrm{e}^{\mathrm{d}\Dec_{z_2}(u_2)})$ for some tangent vectors $u_1$ and $u_2$. Then by using \cite[chapter 5, proposition 2.7]{DoCarmo} which approximates the geodesic distance by the Riemannian distance $h$ on $X$, we obtain that the last term is approximately equal to $\|\mathrm{d}\Dec_{z_1}(u_1)-\mathrm{d}\Dec_{z_2}(u_2)\|_{h(\Dec(z))}$ (to first order) which in turn is equal to $c(z)\|u_1-u_2\|_{g(z)}$ since $\Dec$ is conformal\footnote{Given $(M,g)$ a Riemannian manifold, $\|u\|^2_{g(x)}:=g(x)(u,u)$ (or $u^\mathsf{T}G(x)u$ in matrix form) for any tangent vector $u\in T_xM$.}. Then using the exponential $\exp:T_zZ\to Z$ and the aforementioned proposition once more, we obtain $c(z)d_{\mathsf{geo}}^Z(z_1,z_2)$.
\end{remark}

\begin{remark}
    \label{rem:others}
    The regularization term \eqref{eq:isometric-regularizer} can be used to encourage the decoder $\Dec$ to be \keyword{local isometry}. Using the same function $\Phi$ as before, and computations similar to the ones in proposition \ref{pro:conformal-probabilistic-regularizer} shows that we obtain a regularization term of the form:
    \begin{align}
        \begin{split}
            \label{eq:local-isometric-probabilistic-regularizer}
            \mathcal{L}_{\mathsf{lociso}}(\Dec)=\frac{1}{2m}\Exp_{z\sim\nu}&\left[\Tr R^2(z)\right]\\
            &-\frac{1}{m}\Exp_{z\sim\nu}\left[\Tr R(z)\right]+\frac{1}{2}.
        \end{split}
    \end{align}
    See also \cite{RiemannianGeometricFramework} for a similar approach but that does not make use of neural networks. In \cite{lee2023geometricperspectiveautoencoders}, \keyword{constant} conformal maps were introduced; they correspond to the coordinate-invariant regularization term:
    \begin{equation}
    \label{eqn:conformal-probabilistic-regularizer}
        \mathcal{L}_{\mathsf{conf}}(\Dec)=\frac{m}{2}\frac{\Exp_{z\sim\nu}\left[\Tr R^2(z)\right]}{\Exp_{z\sim\nu}[\Tr R(z)]^2}-\frac{1}{2},
    \end{equation}
    and the associated \keyword{constant} conformal factor then can be seen to be $c\equiv\frac{1}{m}\int_Z\Tr R(z)\ \mathrm{d}\nu$. Notice also the difference with $\widetilde{\mathcal{L}}_{\mathsf{conf}}(\Dec)$ regarding the use of the expectation operator $\Exp_{\nu}$.
\end{remark}

\section{Computing scalar curvature}

In Riemannian geometry, \keyword{scalar curvature} \cite{LeeRiemannian} is the tensor of lowest rank that one can associate to the original Riemann curvature tensor. It is a smooth function $S$ on the manifold (or on the latent space, in case the manifold is embedded like in this article) that measures the curvature at that point, independently from a choice a local chart (or parametrization in our case). Computing different measures of the curvature of the learned manifold gives an \keyword{intrinsic} piece information about this manifold, and is of particular interest in the case of RNAseq data, see \cite{RNAseq-curvature-1} and \cite{RNAseq-curvature-2} for instance. With the help of the regularizer \eqref{eq:nonlinear-conformal-regularizer}, in the case the latent space is of dimension $2$ (which is usually the case for visualization) we are encouraging the decoder to be conformal, meaning that at each latent variables $z\in Z$ we are trying to impose the relationship:
\begin{equation*}
    (\Dec^\star g_{\mathsf{eucl}(n)})(z)=c(z)g_{\mathsf{eucl}(2)}(z),
\end{equation*}
where $g_{\mathsf{eucl}(n)}$ denotes the Euclidean metric on $\setR^n$. Therefore, the scalar curvature on the latent space is the scalar curvature of the metric $cg_{\mathsf{eucl}(m)}$, with $m=2$. However, the change of scalar curvature $S(g)$ (on a manifold of dimension $m$) under a conformal change of metric $g\mapsto\mathrm{e}^{2f}g$ is well known in Riemannian geometry (see \cite{Besse}):
\begin{align*}
    \begin{split}
        S(\mathrm{e}^{2f}&g)=\\
        &\mathrm{e}^{-2f}\big[S(g)-2(m-1)\Delta^gf\\
        &\quad\quad\quad-(m-2)(m-1)\|\mathrm{d}f\|_g^2\big],
    \end{split}
\end{align*}
where $\Delta^g$ denotes the Laplace-Beltrami operator, generalizing the Laplacian to manifolds (see \cite{LeeRiemannian}). In our case $m=2$ and $g=g_{\mathsf{eucl}(2)}$ so we obtain:
\begin{equation}
    S(cg_{\mathsf{eucl}(2)})=-\frac{1}{c}\Delta\log c,
\end{equation}
which is independent of the dimension $n$ of the space in which the data manifold $\mathcal{D}=\Dec(Z)\subset\setR^n$ is embedded. Once the decoder $\Dec$ has been learned during the training phase, we can compute the conformal factor $c$ with the formula \eqref{eq:conformal-factor-formula}: to each input data point corresponds a latent code $z$, and accordingly a value $c(z)$ of the conformal factor at that point. To estimate the scalar curvature of $\mathcal{D}$, a simple approach is then to construct a $k$-nearest neighbors graph from the latent codes, a weight matrix $W$ (and its associated degree matrix $D$) with exponential weights with respect to the Euclidean distances between latent codes, and form the graph Laplacian $L=D-W$. Then if $\mathbf{c}=\{c(z_i)\}_{1\leq i\leq N}$ represents the array of estimated values for the conformal factor, the scalar curvature is given by the array $-\frac{1}{\mathbf{c}}L\log\mathbf{c}$, where the division and the logarithm are component-wise operations. 

\section{Implementation}
\label{sec:implementation}

In order to implement the different regularizers, we use Hutchinson's trace estimator \cite{Hutchinson}. Namely, given a $n\times n$ matrix $A$, the trace $\Tr A$ can be Monte-Carlo estimated thanks to the relation $\Tr A=\Exp_{v\sim\mathcal{N}(0,I_n)}[v^\mathsf{T}Av]$, which has the advantage to only require matrix-vector products. Stacking $n$ samples from a Rademacher distribution instead of using one sample from a $n$-dimensional multivariate Gaussian, one obtains an unbiased estimator with minimal variance (see \cite[proposition 1]{Hutchinson})\footnote{Notice though that Hutchinson's estimator may exhibit a great amount of Monte-Carlo variance. To remedy to this problem, variance reduction techniques can be employed, see this \href{https://www.nowozin.net/sebastian/blog/thoughts-on-trace-estimation-in-deep-learning.html}{web page} for an exhaustive review and insights.} We give an example of this estimator for $\Tr R$ with $R=J^\mathsf{T}J$, $J$ denoting the Jacobian of some map (a decoder in particular). We have the formula
\begin{align*}
    \begin{split}
        \Tr R\approx\frac{1}{N}\sum_{i=1}^N v_i^\mathsf{T}&J^\mathsf{T}Jv_i\\
        &=\frac{1}{N}\sum_{i=1}^N\|Jv_i\|^2=\frac{1}{N}\sum_{i=1}^N\sum_{j=1}^m(Jv_i)_j^2,
    \end{split}
\end{align*}
which we used to implement the regularization term \eqref{eq:nonlinear-conformal-probabilistic-regularizer} with \mbox{PyTorch}\footnote{Matrix-vector products of the form $Jv$ are computed with the help of the function \texttt{torch.func.jvp}, while matrix-vector products such as $J^\mathsf{T}$ (that appear in $R^2)$ are computed with the help of the function \texttt{torch.func.vjp}.}, and where $v_i$ denotes a stacked Rademacher sample as described before. Also, to implement $\Exp_{\nu}$, we take $\nu$ to be the empirical measure.

Furthermore when the full Jacobian of the decoder is needed for a batch of data, we make use of the the \texttt{torch.func.vmap} and  \texttt{torch.func.jacfwd} functions. The optimizer used is the weighted Adam's algorithm, with additions in the case of the Celebrities dataset (see section \ref{ssec:celebrities}).

All code was implemented using PyTorch and was run on standard laptop-grade Intel CPU hardware (i5-11400H), then with a standard NVIDIA GPU hardware (GeForce RTX 3500), as well as on a computing cluster equipped with NVIDIA H100 NVL GPU hardware, giving roughly a tenfold performance increase on the experiments to follow. The full code can be found at this \href{https://github.com/bcoueraud87/geometry_regularized_autoencoders}{URL}.

\begin{remark}
    Concerning reproducibility, although we made sure that all seeds (Python, NumPy, PyTorch and CUDA ones) are set before running the experiments, we couldn't obtain the same values for different runs. According to PyTorch's documentation, this is due to CUDA choosing at runtime what is the best algorithm to run the code on the GPU, and that randomness is involved in this choice. Even with fully deterministic CUDA algorithms we couldn't obtain the same values for different runs. Running multiple times the same experiment and averaging the results was ruled out because it proved to be too computationally intensive. We invite the reader to consult PyTorch's \href{https://docs.pytorch.org/docs/stable/notes/randomness.html}{official documentation} on this matter.
\end{remark}

\section{Experiments}
\label{sec:experiments}

In order to assess the regularization terms $\mathcal{L}_\textsf{lociso}$ \eqref{eq:local-isometric-probabilistic-regularizer} and $\widetilde{\mathcal{L}}_\textsf{conf}$ \eqref{eqn:conformal-probabilistic-regularizer} proposed in this article, we created experiments based on several datasets. A vanilla autoencoder with reconstruction loss $\mathcal{L}_\textsf{recon}$ \eqref{eq:reconstruction-loss} serves as the basis for comparison (with no geometric regularizer), with layers of different types according to the dataset used. Then, the different regularization terms were added to the reconstruction loss $\mathcal{L}_{\mathrm{recon}}$, starting with the well-known $\mathcal{L}_\textsf{gloiso}$ \eqref{eq:global-isometric-regularizer} and then considering $\mathcal{L}_\textsf{lociso}$ and $\widetilde{\mathcal{L}}_\textsf{conf}$.

\subsection{Swiss roll dataset}

The Swiss roll is a classical toy dataset in nonlinear dimensionality reduction. It can be found for example in scikit-learn. In this package it is parametrized as follows: $Z=[\frac{3\pi}{2},\frac{9\pi}{2}]\times[0,21]$, a latent variable is written $z=(\xi,\eta)$, and the parametrization is given by: 
\begin{equation}
    \Dec(\xi,\eta)=\begin{cases}
    x=\xi\cos\xi\\
    y=\eta\\
    z=\xi\sin\xi
    \end{cases}.
\end{equation}
The notations $Z$ and $\Dec$ are precisely chosen so as to give a parallel with the autoencoder structure \eqref{eq:autoencoder} (with $X=\setR^3$, $n=3$ and $m=2$). We are now given samples from this dataset and the aim of the encoder is to obtain a latent space $Z$ (not necessarily the same as the one above), while the decoder will give us a new parametrization of the Swiss roll. Adding the regularization terms means that we are adding constraints on the way the Swiss roll is embedded in $\setR^3$ through its parametrization $\Dec$. Using this parametrization we compute:
\begin{equation*}
    R(\xi,\eta)=J_{\Dec}(\xi,\eta)^\mathsf{T}J_{\Dec}(\xi,\eta)=\begin{bmatrix}
        1+\xi^2 & 0\\
        0 & 1
    \end{bmatrix},
\end{equation*}
therefore the parametrization above is not conformal, and the regularizer we propose will find one. The dataset has been normalized as follows: to each data point the average among all samples is subtracted, and we divide the result by the standard deviation among the samples. The vanilla autoencoder is composed of a sequence of fully connected layers with dimensions $\mathbf{3} \rightarrow \mathbf{50} \rightarrow \mathbf{50} \rightarrow \mathbf{50} \rightarrow \mathbf{2}$ for the encoder, and reversed for the decoder, together with ReLU activation functions. The same architecture is held throughout the experiments, for all regularizers. The standard mean squared error (MSE) loss is used to implement the reconstruction loss. 

\begin{figure}
    \centering
    \includegraphics[scale=0.4]{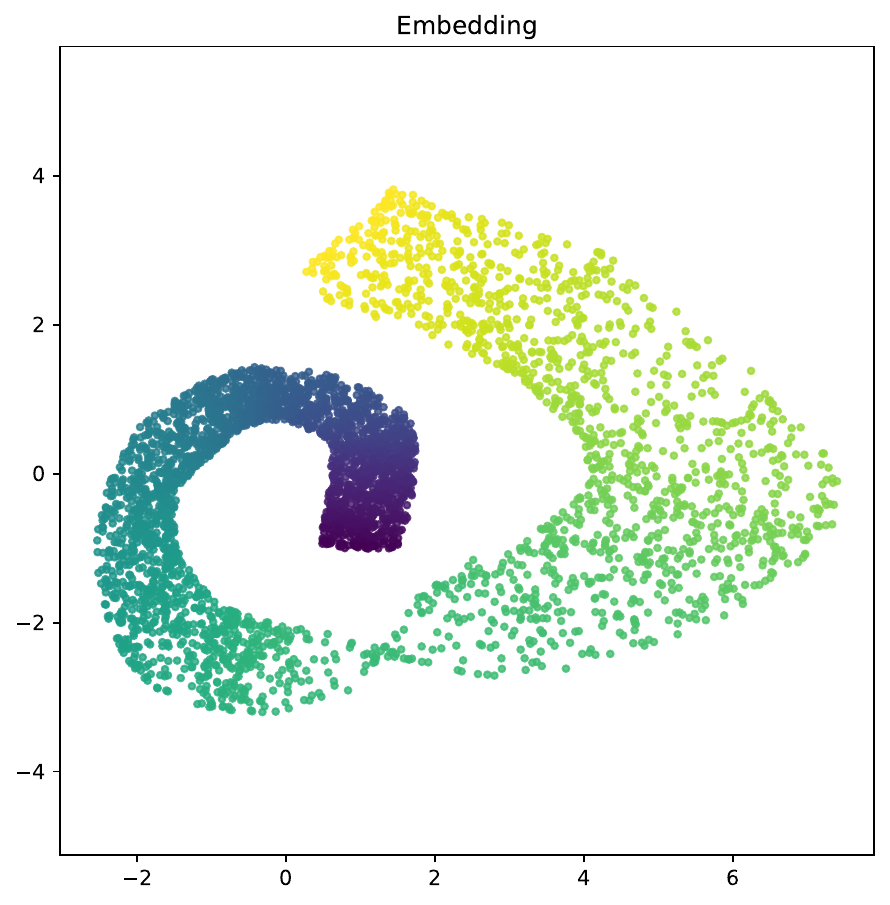}
\end{figure}

\begin{figure}
    \centering
    \includegraphics[scale=0.4]{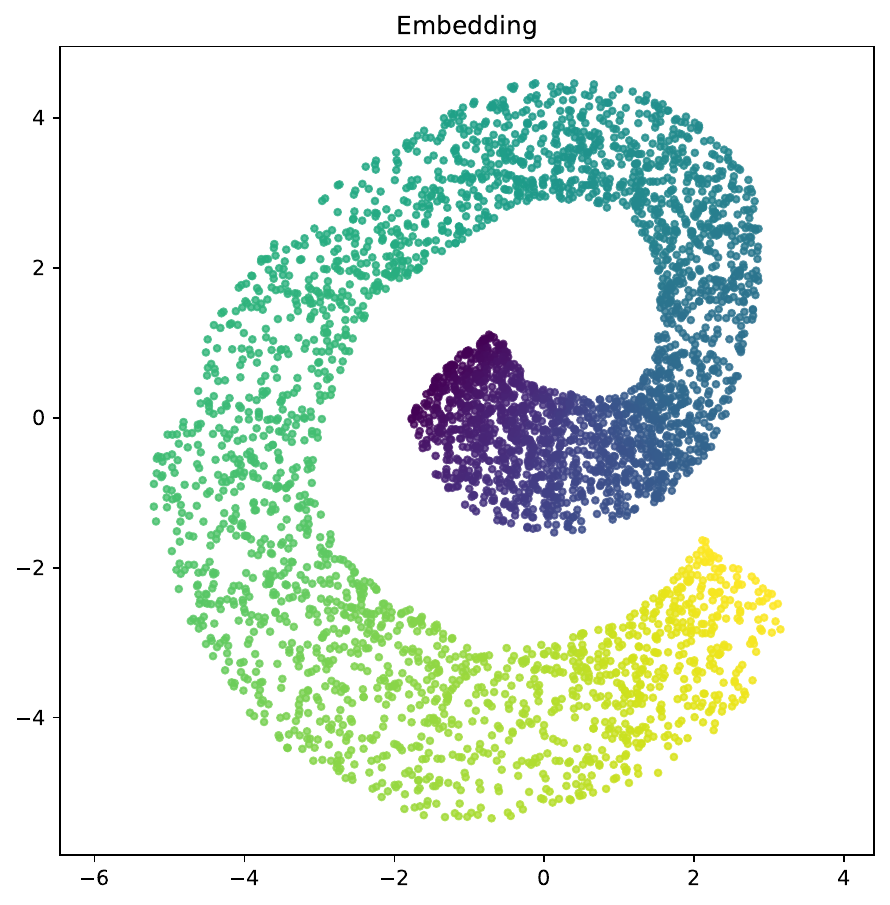}
    \caption{Above, the codes obtained at convergence (200 epochs with batch size 64 and learning rate $10^{-3}$) for the vanilla autoencoder, without any geometric regularization. Below, the ones obtained with nonlinear conformal regularization.}
\end{figure}

\begin{figure}
    \centering
    \includegraphics[scale=0.4]{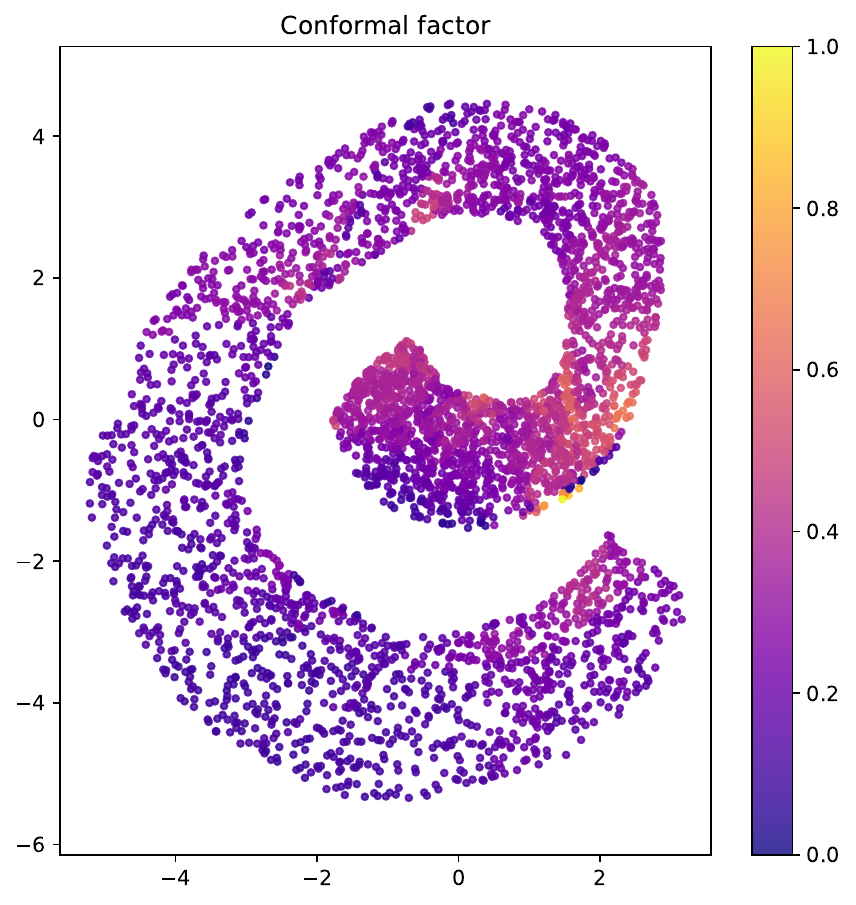}
    \caption{Normalized conformal factor obtained with the $\widetilde{\mathcal{L}}_\textsf{conf}$ regularizer, added to the reconstruction loss $\mathcal{L}_{\mathrm{recon}}$ of the vanilla autoencoder. Stretching is the greatest inside the roll.}
\end{figure}

\begin{figure}
    \centering
    \includegraphics[scale=0.4]{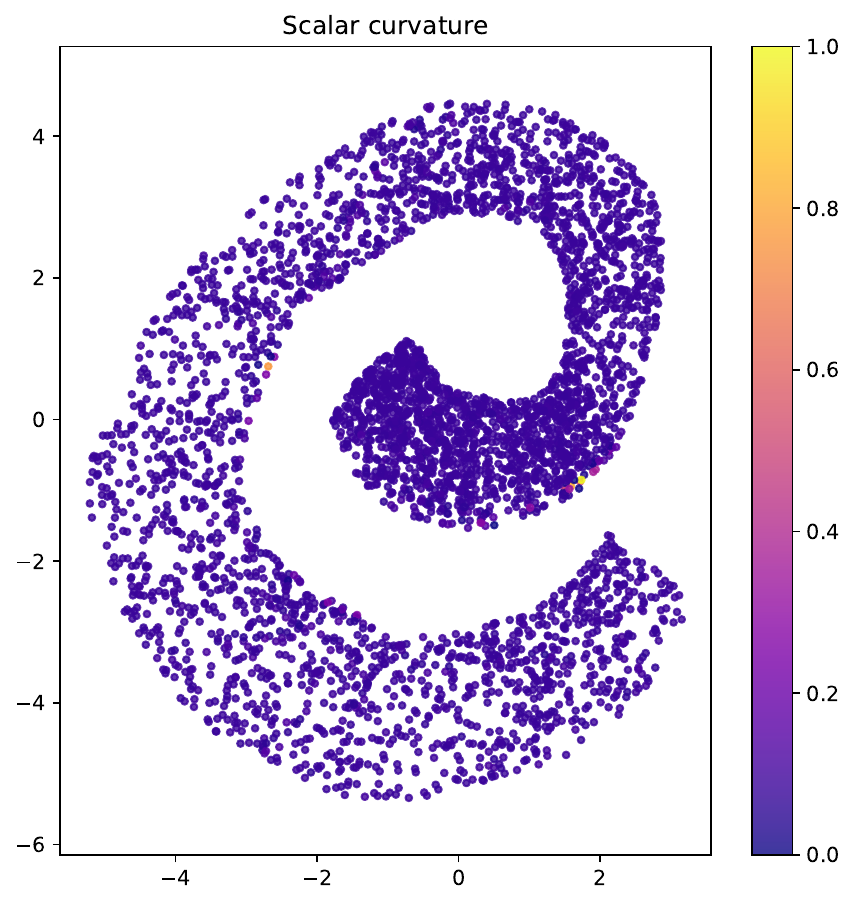}
    \caption{Normalized scalar curvature obtained with the $\widetilde{\mathcal{L}}_\textsf{conf}$ regularizer, added to the reconstruction loss $\mathcal{L}_{\mathrm{recon}}$ of the vanilla autoencoder, by computing a graph Laplacian on the latent space. We recover the null scalar curvature of the Swiss roll, except at a few points on the boundary.}
    \label{fig:conformal-factor}
\end{figure}

The scalar curvature of the Swiss roll, which is twice its Gaussian curvature, can be computed from the parametrization above and we find $S\equiv 0$ as expected as it is an embedding of a plane in $\setR^3$ (see \cite{Pressley} for how to compute the Gaussian curvature from the first and second fundamental forms). The vanilla autoencoder, together with the conformal regularizer $\widetilde{\mathcal{L}}_\textsf{conf}$ and the computation of the graph Laplacian on the latent space allows to recover this fact as shown by figure \ref{fig:conformal-factor}. 

\begin{figure}
    \includegraphics[scale=0.4]{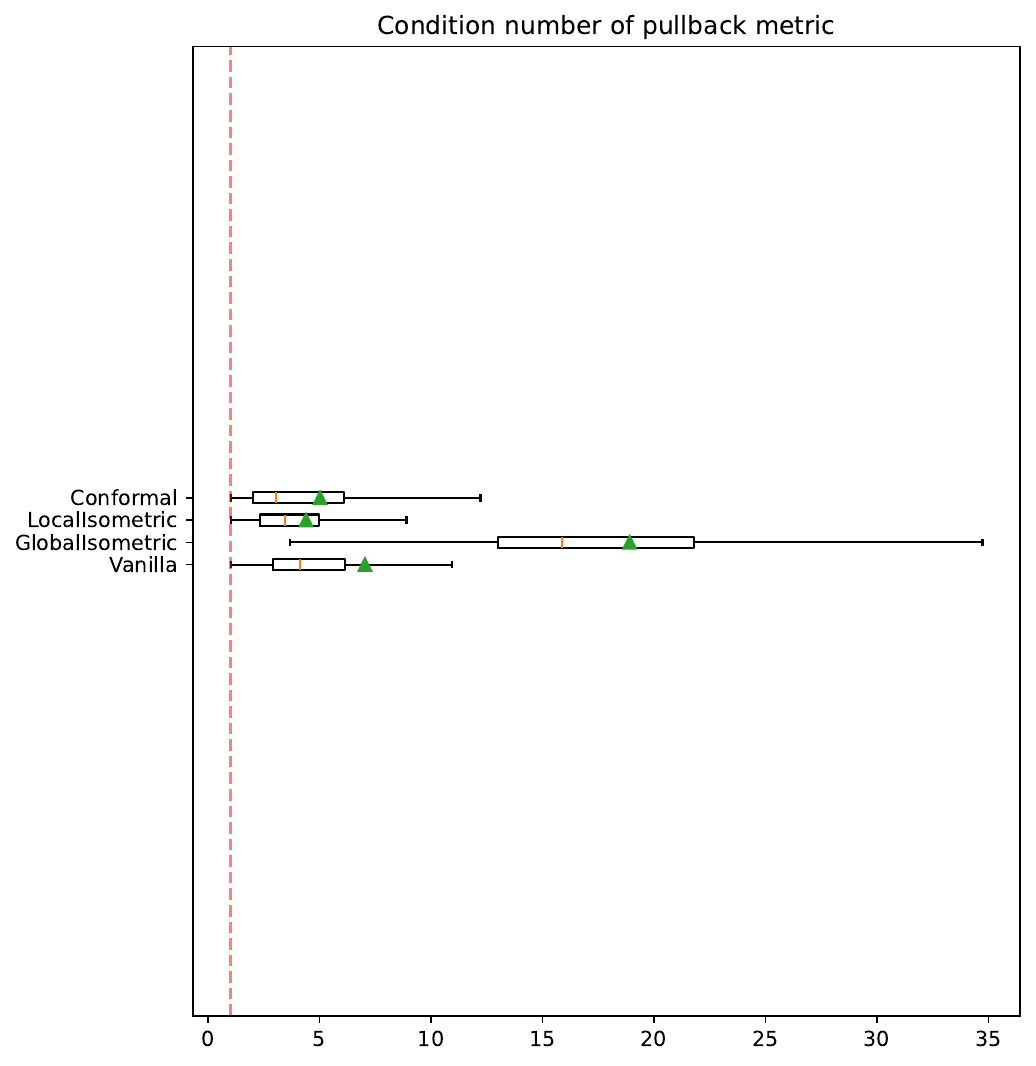}
    \caption{Spatial distribution of the condition numbers of the decoder's Jacobian (first plot) and of the pullback metric (second plot). A vertical dashed red line highlights the line on which the condition number is exactly one, and an green triangle indicates the distribution mean.}
    \label{fig:kappa-distributions}
\end{figure}

On the validation dataset we also measured the condition number (with respect to the $L^2$ norm) $\kappa_\textsf{jac}$ of the Jacobian of the decoder, $J_{\Dec}(z)$, and the condition number $\kappa_\textsf{pbm}$ of the pull-back metric of the decoder, $J_{\Dec}(z)^{\mathsf{T}}J_{\Dec}(z)$ with $z\in Z\subset\setR^2$. The first measure shows how the linearized decoder around one latent code amplifies variations in its input (latent variables), and the second summarizes the pullback metric, specifically the anisotropy of the pullback metric, or how far the decoder is from being a local isometry. These condition numbers are computed with \texttt{torch.linalg.cond} at each validation point in the latent space. Their spatial distributions are visualized in figure \ref{fig:kappa-distributions} and summarized in the table \ref{tab:results-swiss-roll}, for all geometric regularizers. Although no experiment repetition and results averaging has been performed, it has been observed clearly that nonlinear conformal regularization (and local isometric regularization) consistently learn a better conditioned and anisotropic decoder.  

\begin{figure}
    \centering
    \includegraphics[scale=0.4]{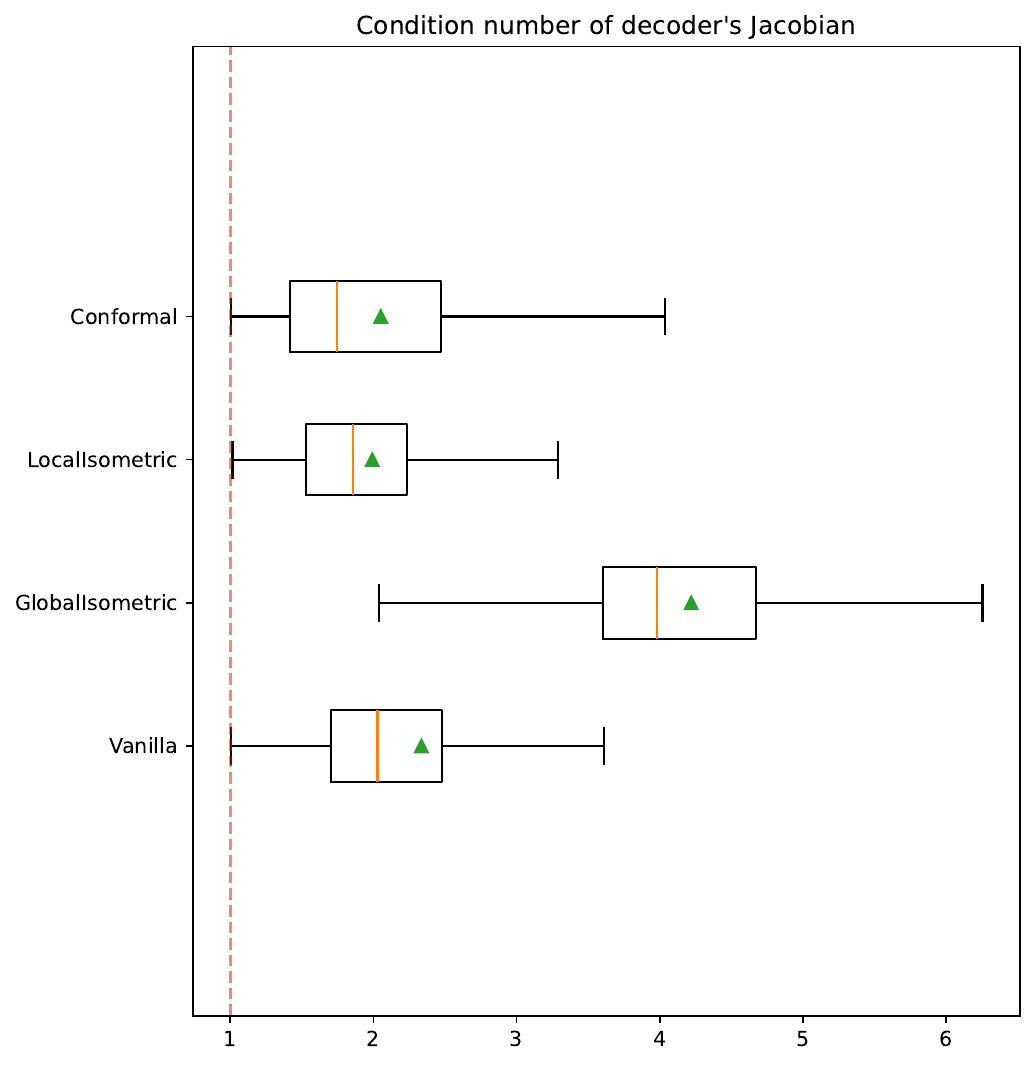}
\end{figure}

\begin{table}
    \centering
    \begin{tabular}{|c||c|c|}
    \hline
    & $\mathcal{L}_{\mathsf{recon}}$ & $\mathcal{L}_{\mathsf{globiso}}$ \\ \hline\hline
    $\kappa_\mathsf{jac}$ & $2.33\pm1.26$ & $4.22\pm1.05$ \\ \hline
    $\kappa_\mathsf{pbm}$ & $7.03\pm13.32$ & $18.91\pm12.54$ \\ \hline
    & $\mathcal{L}_{\mathsf{lociso}}$ & $\widetilde{\mathcal{L}}_{\mathsf{conf}}$ \\ \hline\hline
    $\kappa_\mathsf{jac}$ & $1.99\pm0.65$ & $2.05\pm0.90$ \\ \hline
    $\kappa_\mathsf{pbm}$ & $4.39\pm3.23$ & $5.02\pm5.29$ \\ \hline
    \end{tabular}
    \caption{Results for the conditions numbers $\kappa_\mathsf{jac}$ of the decoder's Jacobian, and $\kappa_\mathsf{pbm}$ of the pullback metric, respectively (mean $\pm$ standard deviation), across the whole latent space. The vanilla autoencoder together $\mathcal{L}_{\mathsf{lociso}}$ or $\widetilde{\mathcal{L}}_{\mathsf{conf}}$ consistently outperform the vanilla autoencoder alone or with the regularizer $\mathcal{L}_{\mathsf{globiso}}$, and we observed that the difference is even more important upon increasing the number of samples in Hutchinson's estimator.}
    \label{tab:results-swiss-roll}
\end{table} 

An important hyperparameter to consider when one wants to quantitatively assess the benefits of a geometric regularizer is the intensity with which the constraint is imposed in the total loss. One rule of thumb is to first adjust it so that the regularizer values times this intensity are in the same range as the values of the reconstruction loss $\mathcal{L}_{\mathsf{recon}}$. Then, from that stage, if we increase this intensity it may happen that the reconstruction loss does not decrease anymore, while decreasing the intensity will accordingly decrease the desired effect of the geometric regularizer. Finding a good balance is a challenge, especially because the regularizer values change during the training process.

\subsection{Celebrities dataset}
\label{ssec:celebrities}

The Celebrities dataset (also widely known as CelebA, see \cite{CelebA}) is a well-known dataset that comprises more than 200,000 faces of celebrities (with 40 annotated attributes), each of size $178\times 218$ pixels. With the help of the \texttt{torchvision.transforms} submodule, each image in the dataset is center cropped to $150\times 150$ pixels, resized to $32\times32$ pixels, and normalized per channel so that all values are in $[-1,1]$. Therefore, with regards to the structure \eqref{eq:autoencoder}, $n=3072$ and $m=2$. For this experiment the vanilla autoencoder uses convolutional layers with kernel size $3$, stride $2$ and padding $1$, together with leaky ReLU activation functions, except for the last layer of the decoder which uses an hyperbolic tangent in order to normalize the values in $[-1,1]$ and recreate an image. There is also a fully connected layer after the last convolution layer to map the last feature map onto the latent space. The dimensions are as follows: $\mathbf{3\times 32\times 32} \rightarrow \mathbf{32\times 16\times 16} \rightarrow \mathbf{64\times 8\times 8} \rightarrow \mathbf{128\times 4\times 4} \xrightarrow{\text{flatten}} \mathbf{2048} \rightarrow \mathbf{2}$. In order to reduce overfitting, for this dataset we used weight decay and a learning rate scheduler that reduces the learning rate upon encountering a plateau (starting with a learning rate of $10^{-4}$, with batch size 64, attaining convergence at 100 epochs). Note that since the data set is high dimensional and the compression is maximal (latent space has dimension 2), it is very difficult to reconstruct images and the reconstruction loss stabilizes around $0.1$ only. The table \ref{tab:results-celeba} shows once more that the local isometry and nonlinear conformal regularizers added to the vanilla autoencoder learn better conditioned decoders than the global isometry regularizer.

\begin{table}
    \centering
    \begin{tabular}{|c||c|c|}
    \hline
    & $\mathcal{L}_{\mathsf{recon}}$ & $\mathcal{L}_{\mathsf{globiso}}$ \\ \hline\hline
    $\kappa_\mathsf{jac}$ & $1.91\pm0.58$ & $33.84\pm10.10$ \\ \hline
    $\kappa_\mathsf{pbm}$ & $3.97\pm2.93$ & $1246.92\pm651.46$ \\ \hline
    & $\mathcal{L}_{\mathsf{lociso}}$ & $\widetilde{\mathcal{L}}_{\mathsf{conf}}$ \\ \hline\hline
    $\kappa_\mathsf{jac}$ & $2.06\pm0.59$ & $1.74\pm0.44$ \\ \hline
    $\kappa_\mathsf{pbm}$ & $4.59\pm3.37$ & $3.23\pm1.65$ \\ \hline
    \end{tabular}
    \caption{Results for $\kappa_\mathsf{jac}$ and $\kappa_\mathsf{pbm}$ on the CelebA dataset: the vanilla autoencoder together $\mathcal{L}_{\mathsf{lociso}}$ or $\widetilde{\mathcal{L}}_{\mathsf{conf}}$ consistently outperform the vanilla autoencoder alone or with the regularizer $\mathcal{L}_{\mathsf{globiso}}$.}
    \label{tab:results-celeba}
\end{table} 

\section{Conclusion}

In this article, we applied a new type of \keyword{geometric regularization} to decoders, namely \keyword{nonlinear conformal regularization}. \keyword{Conformal maps} do not exactly preserve distances, only up to a certain factor that varies with the codes in the latent space, called the \keyword{conformal factor}, which allows to make an assessment on the local deformation that occurs upon reconstructing the observed data from latent codes. The regularizer is formulated in a \keyword{coordinate-invariant} way, in the spirit of \cite{RiemannianGeometricFramework}, and is implemented as a \keyword{Hutchinson's Monte-Carlo estimator}, whose quality increases with the number of stacked Rademacher samples. The architecture proposed automatically learns the conformal factor as well as the \keyword{scalar curvature} of the manifold after construction of a graph Laplacian on the latent space. In addition, it is more flexible than a global isometry regularizer in the sense that \keyword{local variations} of the decoder are allowed, and the regularizers allow the architecture to learn \keyword{better conditioned} decoders.

This work opens further questions on geometric regularization in the context of autoencoders. One venue of research would be to integrate this new geometric regularizer into a variational autoencoder (VAE), and study the interplay between the geometry brought back onto the latent space with the help of the pullback metric, and the information geometry of the latent space as given bye the VAE (several works are already investigating this direction, such as \cite{CA22} and \cite{KEAH20}), and how conformal maps are beneficial in that case. Moreover, VAEs tend to use less samples than standard autoencoders, which would be beneficial to study RNAseq datasets with the presented framework, possibly opening the way to make statements based on stretched distances and scalar curvature as visualized in the latent space. Also, it would be interesting to study variance reduction techniques for the conformal regularizer, with the aim of improving both the condition number of the decoder's Jacobian and the condition number of the pullback metric.\\

\textbf{Acknowledgments:} The author would like to thank Enikő Regényi for her explanations regarding RNAseq data and for the preparation of experiments regarding several RNAseq datasets (unpublished).

\bibliography{article}

\end{document}